\documentclass[]{atom_air}

\newif\ifshowbackmatter
\showbackmattertrue

\newif\ifshowappendices
\showappendicesfalse

\usepackage[toc,page,header]{appendix}
\usepackage{titletoc}

\usepackage{graphicx}%
\usepackage{multirow}%
\usepackage{amsmath,amssymb,amsfonts}%
\usepackage{amsthm}%
\usepackage{mathrsfs}%
\usepackage{xcolor}%
\usepackage{textcomp}%
\usepackage{manyfoot}%
\usepackage{tabularx}
\usepackage{colortbl}
\usepackage{array}
\usepackage{makecell}
\usepackage{ragged2e}
\usepackage{booktabs}%
\usepackage{algorithm}%
\usepackage{algorithmicx}%
\usepackage{algpseudocode}%
\usepackage{listings}%
\usepackage{pdflscape}
\usepackage{seqsplit}
\usepackage{longtable}
\usepackage[expansion=false]{microtype}
\usepackage{eso-pic}
\usepackage{transparent}

\newif\ifconfidentialcopy
\confidentialcopyfalse
\ifconfidentialcopy
\AddToShipoutPictureFG{%
  \AtPageCenter{%
    \makebox(0,0){%
      \rotatebox{45}{%
        \texttransparent{0.10}{%
          \color{black}\sffamily\bfseries\fontsize{70}{70}\selectfont
          CONFIDENTIAL}%
      }%
    }%
  }%
}
\fi

\title{\aiphergradienttext{PharmAgents2}: Towards Autonomous End-to-End Drug Discovery}

\renewcommand{\authorfont}{\fontfamily{qpl}\fontseries{m}\fontshape{n}\fontsize{11}{14}\selectfont\color{aipherInk}}
\renewcommand{\authorformat}[2][]{{\authorfont #2$^{#1}$}}
\author{AIPher Technology\\[0.5em]Institute for AI Industry Research (AIR), Tsinghua University}

\abstract{
Drug discovery is a complex scientific process that demands continuous decision-making across a vast research space. Making high-quality scientific decisions and exploring this space effectively are two central challenges in this process. To address these challenges, we introduce {\bfseries PharmAgents2}, an autonomous system for end-to-end drug discovery. To support high-quality scientific decisions, we propose Paper2Skill, a framework that extracts transferable scientific decision knowledge from publicly available sources. Together with expert-developed skills, this knowledge forms the foundation of PharmAgents2's decision-making capabilities. To enable organized exploration across a broad research space, we develop a Self-Scaling Discovery framework that scales exploration through parallel agents inheriting research context and decision-making capabilities, spanning diverse research spaces, including those of targets, scientific hypotheses, and discovery strategies. We further introduce \mbox{\bfseries PharmAgents-Bench v0.1}, which evaluates autonomous discovery at two complementary levels: the quality of individual scientific decisions and the ability to carry a research task through to completion. It includes 121 questions across nine decision scenarios and three end-to-end discovery tasks assessed using expert-defined process rubrics. PharmAgents2 outperforms the evaluated baselines in most decision categories and achieves the highest process scores on all three end-to-end tasks. Across three real-world research scenarios, PharmAgents2 autonomously generated discoveries through end-to-end workflows, with key findings subsequently validated in human-executed experiments. For inhibitor discovery against the important antibacterial target LpxH, it autonomously hypothesized a previously unreported allosteric site and screened compounds against this proposed site, identifying a novel inhibitor scaffold with experimentally confirmed inhibitory activity and binding. For selective inhibitor discovery against JAK1, a therapeutic target in inflammatory and autoimmune diseases, it autonomously investigated potential strategies, prioritized a covalent inhibition strategy, and nominated compounds with experimentally demonstrated activity. For target exploration starting from a known bioactive molecule, Self-Scaling Discovery enabled systematic exploration across the proteome and identified a new binding target, with the interaction subsequently confirmed experimentally. PharmAgents2 thus represents a prototype of a virtual pharmaceutical company, marking a step toward autonomous end-to-end drug discovery.

}

\newcommand{\methodabbr}{PharmAgents2}
\newcommand{\paperoskill}{Paper2Skill}

\newtcolorbox{flowcaption}{blanker,breakable,
  before skip=0pt,after skip=0pt,
  title after break={Figure~\thefigure\ (continued)},
  fonttitle=\small\sffamily\bfseries,coltitle=aipherNavy,
  pad before break=12pt,pad after break=0pt,lines before break=2,
  overlay first and middle={%
    \node[anchor=south east,inner sep=0pt,yshift=1pt,
      font=\footnotesize\itshape,text=aipherMuted]
      at (frame.south east) {(continued on next page)};
  }}

\begin{document}
\maketitle


\clearpage
\section{Introduction}
\label{sec:intro}

Drug discovery is central to addressing unmet medical needs and improving human health. In recent years, AI has been reshaping this field, with the potential to fundamentally transform how new therapeutics are discovered. A growing range of specialized models has enabled advances and improved efficiency across individual research tasks, exemplified by AlphaFold in protein structure prediction~\citep{jumper2021alphafold}, RFdiffusion in de novo protein design~\citep{watson2023rfdiffusion}, and PocketXMol in structure-based molecular design~\citep{peng2026pocketxmol}. These advances strengthen individual components of drug discovery, but their contribution to the overall process depends on effective coordination across interdependent research stages. Organizing and advancing this process continues to rely on human direction and judgment. Extending these capabilities from individual tasks to autonomous research across the full discovery process remains a central challenge.

Recent advances in AI agents have opened a new avenue for autonomously conducting complex scientific research. This direction has motivated a growing body of work in scientific discovery and biomedicine. Co-Scientist generates and iteratively refines novel scientific hypotheses~\citep{gottweis2026coscientist}. Biomni provides a general-purpose biomedical agent that automates diverse research tasks, reducing the burden of repetitive and fragmented research workflows~\citep{huang2026biomni}. The Virtual Biotech coordinates specialist agents to integrate evidence across disciplines and provide analysis and decision support for therapeutic development~\citep{zhang2026virtualbiotech}. While these systems automate research tasks and support scientific decisions, a gap remains between these capabilities and autonomous end-to-end drug discovery, from an initial research objective to the delivery of molecular candidates ready for experimental testing.

We view drug discovery as a continuous process of scientific decision making across a vast space of targets, mechanisms, molecules, and research strategies. Each decision shapes the evidence gathered and the options available at subsequent stages, so a poorly supported choice can divert an entire investigation. Advancing a research direction in depth therefore depends on maintaining decision quality throughout the process. This requires an agent to assess the available evidence, determine appropriate next steps, and revise its approach as results emerge, even when experimental feedback is delayed or incomplete. At the same time, a carefully pursued direction covers only a small part of the possible discovery space. Research breadth is therefore equally necessary: alternative hypotheses must be investigated and compared to identify promising directions that an early commitment might otherwise leave unexplored. Because each investigation can require substantial literature analysis and scientific computation, achieving this breadth calls for the capacity to extend research effort across many directions.

\begin{figure}[tp]
\captionlistentry[figure]{Autonomous end-to-end drug discovery with \methodabbr{}.}
\label{fig:positioning}
\centering
\includegraphics[width=\linewidth,height=0.84\textheight,keepaspectratio]{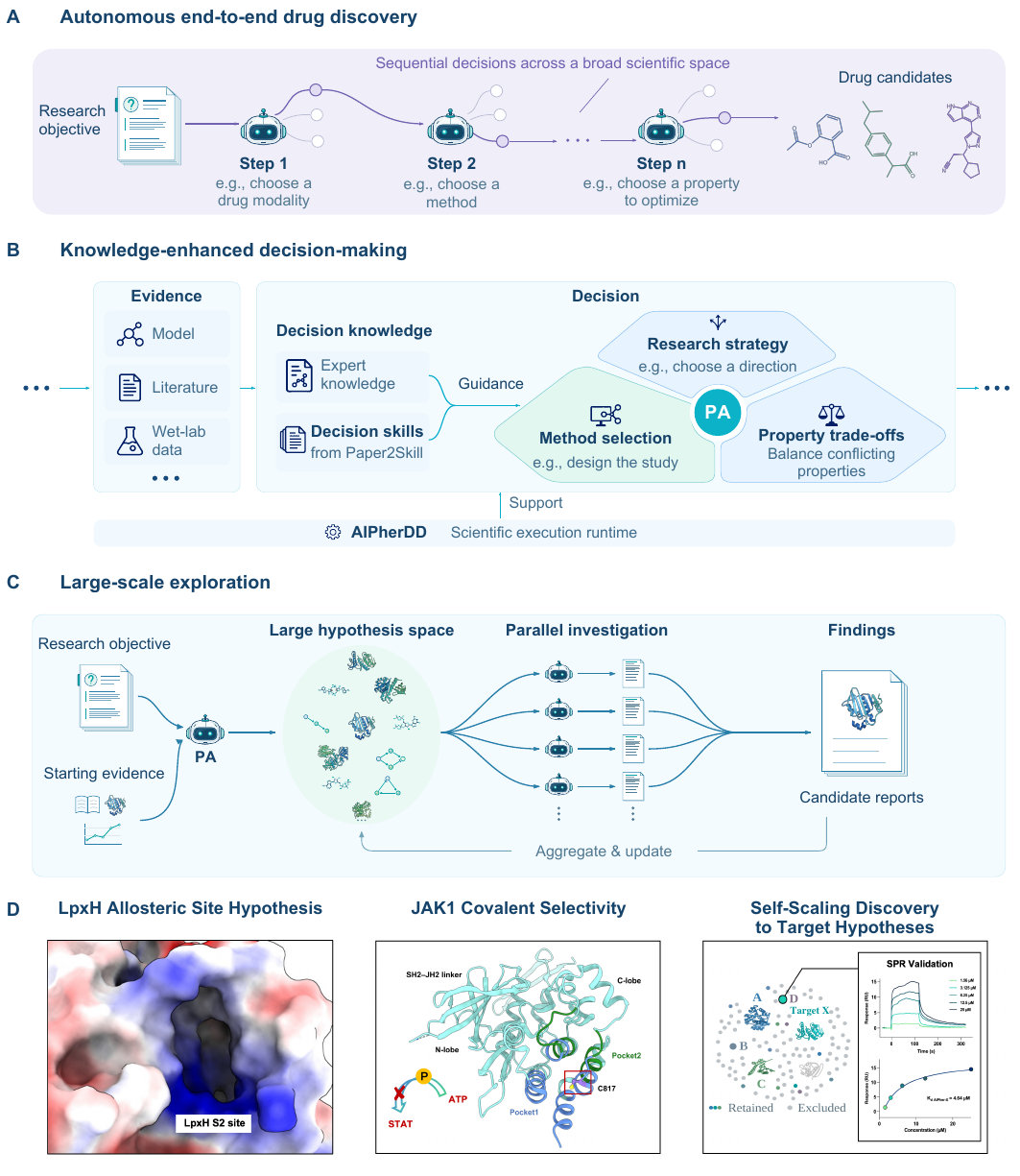}\par
\vskip\abovecaptionskip
\raggedright\small
\setlength{\parskip}{0pt}
\noindent{\color{aipherNavy}\textbf{Figure~\thefigure}}\quad
\textbf{Autonomous end-to-end drug discovery with \methodabbr{}.}
  \textbf{A,} We view drug discovery as a continuous decision-making process
  across a broad scientific space, guided by a research objective and directed
  toward candidate delivery.
  \textbf{B, Depth:} the decision system brings together task-specific evidence
  and knowledge from experts and \paperoskill{} to determine what to investigate
  next, select appropriate methods, and weigh competing molecular properties
  against project requirements. Computational findings inform subsequent
  decisions, allowing the system to revise its research strategy as evidence
  accumulates. AIPherDD provides the scientific models and tools that support
  this execution.
\par\hfill{\footnotesize\itshape\color{aipherMuted}(continued on next page)}
\end{figure}

To address these challenges, we introduce \textbf{PharmAgents2}, an autonomous system for end-to-end drug discovery (Figure~\ref{fig:positioning}). To support high-quality scientific decisions, we develop a knowledge-enhanced decision system built around a scientific decision knowledge base. This knowledge base brings together skills distilled from human expert experience and skills derived from scientific publications through Paper2Skill, a framework we develop to transform published research into transferable decision knowledge. Paper2Skill synthesizes methodologically related studies, capturing the connections between research objectives, methodological choices, and the conditions under which those choices are appropriate. The resulting skills provide reusable guidance that enables the agent to select suitable methods and adapt research procedures to the scientific problem at hand. A supporting computational platform, AIPherDD, enables the execution of scientific models and tools~\citep{gao2023drugclip,fu2024admetlab,swanson2024admetai,gao2026genomewide}, providing task-specific evidence that the system evaluates using its decision knowledge to determine how the investigation should proceed. To enable scientific exploration across a broad research space, we develop Self-Scaling Discovery, a framework for organizing and scaling autonomous scientific investigations. The framework structures the research space into focused questions and coordinates parallel agents to investigate them within a shared research objective. Each agent inherits the relevant research context, allowing it to pursue a focused investigation while retaining its connection to the broader discovery task. Findings across investigations are brought together to update the research space, refine priorities, and guide subsequent rounds of exploration. This organization enables systematic exploration across diverse targets, mechanisms, and discovery strategies while maintaining a coherent research direction as the scale of investigation grows. Together, these capabilities enable PharmAgents2 to advance research from an initial discovery objective to experimentally testable candidate delivery through informed scientific decisions and systematic exploration.

\begingroup
\clubpenalty=10000
Recent benchmarks evaluate biomedical agents from several perspectives: BiomniBench-DA assesses the quality of biomedical analyses using expert-designed rubrics~\citep{qu2026biomnibench}, DrugDiscoveryBench measures computational task accuracy in early-stage drug discovery~\citep{akyurek2026drugdiscoverybench}, and SMDD-Bench evaluates small-molecule design capabilities~\citep{han2026smddbench}. However, these evaluations leave open whether an agent can autonomously carry a drug discovery objective through to experimentally testable candidate delivery. To address this gap, we introduce \textbf{PharmAgents-Bench v0.1}, which evaluates scientific decision quality and end-to-end research execution at two complementary levels. The first track assesses scientific decisions through 121 questions across nine drug discovery scenarios, with reference outcomes grounded in published evidence. The second evaluates autonomous research and candidate delivery through three tasks derived from real drug discovery projects, using expert-defined criteria to assess the quality of the research process. These benchmark results demonstrate that, compared with the evaluated baselines, PharmAgents2 makes higher-quality scientific decisions and conducts end-to-end investigations that more closely align with expert-defined expectations.
\par
\endgroup

We validate PharmAgents2's autonomous drug discovery capabilities in three real-world research scenarios. Starting from research objectives, PharmAgents2 autonomously conducts the investigations and delivers multiple key findings, which are subsequently validated in human-executed wet-lab experiments. To discover novel inhibitor scaffolds for the antibacterial target LpxH~\citep{young2013lpxh,cho2020lpxh,huseby2024lpxh,zhang2026lpxh}, PharmAgents2 hypothesized a previously unreported allosteric site and developed a screening strategy targeting this proposed site. This investigation yielded AIPher-L1, representing a novel inhibitor scaffold, with subsequent experiments confirming its inhibitory activity against LpxH and its binding to the target. Further functional assays provided preliminary support for the proposed allosteric mechanism. To discover selective inhibitors of JAK1, a therapeutic target in inflammatory and autoimmune diseases, PharmAgents2 prioritized a covalent inhibition strategy~\citep{west2018pf04965842,kavanagh2022jak1,caveney2023jak,natamycin2025jak1}, investigated potential binding sites, and nominated candidate compounds for experimental testing. Subsequent cellular assays identified AIPher-J1, which exhibited inhibitory activity comparable to the positive controls in two cytokine-signaling pathways at 10\,$\mu$M, alongside three additional compounds showing greater than 50\% inhibition in both pathways. To identify additional targets of the known bioactive molecule AIPher-X, Self-Scaling Discovery organized systematic investigations across a reference space of 2,607 proteins and identified a previously unrecognized binding target. Subsequent SPR experiments confirmed binding between AIPher-X and its top-ranked target, with a dissociation constant of 4.64\,$\mu$M. Together, these studies demonstrate PharmAgents2's ability to translate research objectives into experimentally supported discoveries through autonomous end-to-end investigations.

PharmAgents2 thus represents a prototype of a virtual pharmaceutical company. Through continuous scientific decision making across a broad research space, it autonomously advances drug discovery from an initial research objective to molecular candidates ready for experimental testing. This work points toward a new paradigm of pharmaceutical innovation in which autonomous AI systems expand the scale and scope of scientific discovery.

\begin{figure}[t]
\raggedright\small
\setlength{\parskip}{0pt}
\noindent{\color{aipherNavy}\textbf{Figure~\ref{fig:positioning} (continued)}}\quad
\textbf{C, Breadth:} Self-Scaling Discovery organizes the hypothesis space into
  research directions and forks parallel agents that share the research context
  and decision capabilities. Each agent investigates an alternative hypothesis
  or strategy, while findings across branches are integrated to compare
  possibilities and guide further exploration. This process enables the system
  to pursue multiple plausible directions and adjust their priority as the
  investigation develops.
  \textbf{D,} Autonomous investigations followed by human-executed wet-lab validation.
  PharmAgents2 hypothesized a previously unreported allosteric site in LpxH
  and screened candidate inhibitors, pursued a covalent strategy toward selective
  JAK1 inhibition and nominated candidate compounds, and performed large-scale
  secondary-target identification for a bioactive molecule.
\end{figure}

\section{PharmAgents2}
\label{sec:method}

\methodabbr{} conducts autonomous drug discovery from a research objective,
project requirements, and available starting information. The agent develops
a research strategy, gathers evidence through literature investigation and
scientific computation, and revises its plan as findings accumulate.
Scientific decisions are guided by expert-authored skills and knowledge
acquired from publications through \paperoskill{}. Self-Scaling Discovery
extends this process across alternative hypotheses through parallel
investigations that inherit the agent's research context and decision
capabilities. AIPherDD provides access to the scientific models and tools
used in these investigations. Before delivery, an independent reviewer agent
examines the investigation and its supporting evidence, returning feedback
for revision when needed. The system delivers candidate lists alongside
research reports documenting its analyses, supporting evidence, and
recommendations for experimental follow-up.
\ifshowbackmatter
\ifshowappendices
The detailed execution architecture is provided in Appendix~\ref{app:impl}.
\fi
\fi

\subsection{Knowledge-Guided Scientific Decision Making}
\label{subsec:skills}

\methodabbr{} supports scientific decision making by integrating task-specific
evidence with reusable decision knowledge. Evidence from literature, data,
and scientific computations establishes what is known about the current
research problem. Decision knowledge guides the agent in assessing research
strategies, choosing appropriate methods, interpreting results, and
determining what to investigate next. These decisions can initiate further
evidence gathering or lead to revisions of the research strategy as the
investigation progresses.

The decision knowledge is organized as a library combining
expert-authored skills with skills acquired from scientific publications
through \paperoskill{}. Relevant skills are retrieved during both planning
and execution, providing methodological guidance for the questions
encountered at each stage. The agent applies this guidance in the context
of the current objective and observations, adapting its choices to the
conditions of the investigation.

\begin{figure}[!t]
  \centering
  \includegraphics[width=\linewidth]{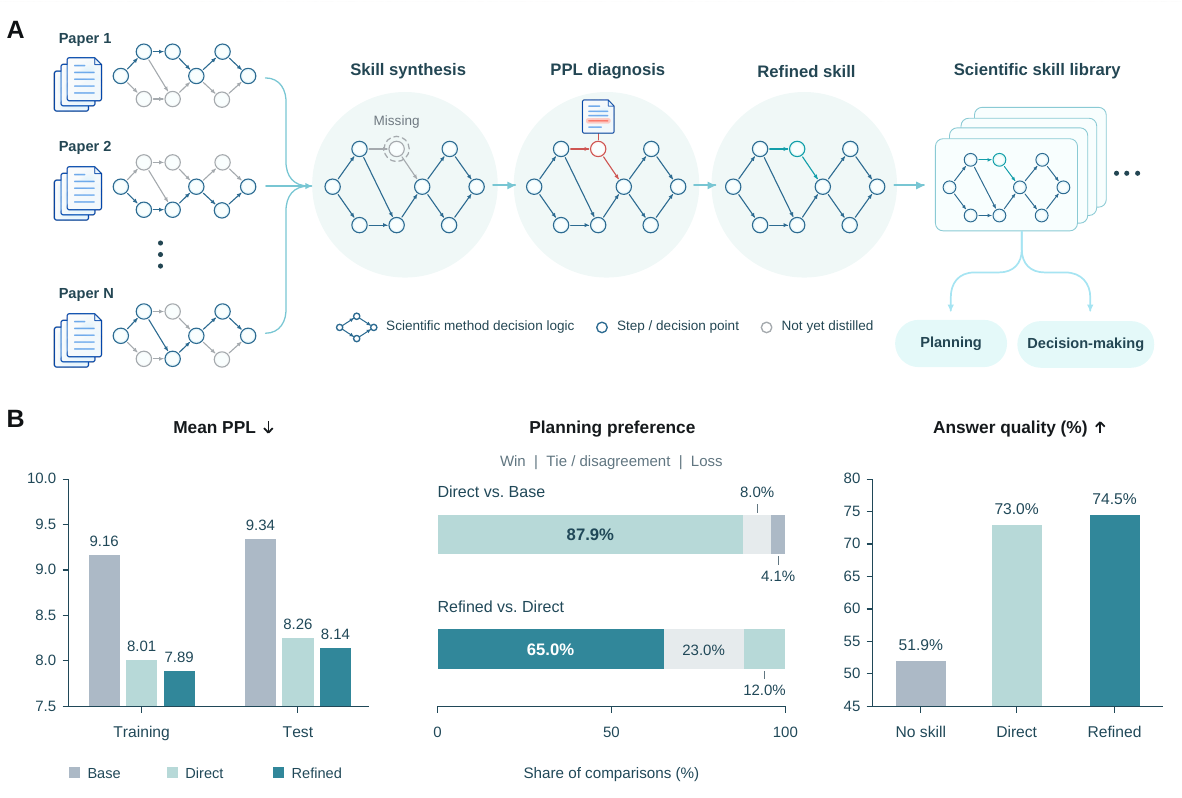}
  \caption{\textbf{Paper2Skill: skill acquisition and evaluation.}
  \textbf{A,} Paper2Skill synthesizes reusable scientific decision knowledge
  from papers that share a methodological workflow. Each paper contributes
  research steps, decision points, and the conditions linking them; these
  are combined into a skill that captures shared procedures and alternative
  research paths. Nodes and connecting arrows schematically represent this
  methodological decision logic, with gray nodes indicating knowledge not
  yet captured in the synthesized skill. For refinement, a fixed language
  model evaluates how well the research context and skill support prediction
  of methodological passages from the source papers. Passages with high
  residual prediction loss are flagged for closer inspection, guiding the
  synthesis model to revisit the source context and supplement omitted
  conditions or methodological relationships. The refined skills are
  collected into a scientific skill library to support planning and
  decision-making.
  \textbf{B,} Evaluation across ten topics. Direct and Refined denote initial
  distillation and perplexity (PPL)-guided refinement, respectively.
  Left: mean PPL on training and test passages, with lower values indicating
  better prediction. Center: paired planning preferences (800 cases per
  comparison). From left to right, segments show wins, ties or order
  disagreements, and losses for the first-named method. Wins and losses
  require agreement in both presentation orders.
  Right: answer quality, measured as the percentage of paper-specific binary
  criteria satisfied. PPL and answer-quality scores use equal topic weights.}
  \label{fig:system}
\end{figure}

\paperoskill{} acquires reusable decision knowledge by synthesizing groups
of papers that share a methodological workflow (Figure~\ref{fig:system}A).
It considers both the research needs addressed by the studies and the
methods they used, preserving the connection between methodological choices
and their scientific context. The resulting skills describe shared research
procedures, relationships between steps, and conditional alternatives. This
representation allows the agent to identify applicable methods and adapt
the research path when a new task differs from the studies from which the
knowledge was acquired.

We further refine the extracted knowledge through perplexity (PPL)-guided knowledge
completion. A fixed language model measures how well the research context
and synthesized skill support prediction of methodological passages from
the construction papers. Passages with high residual prediction loss
identify material for closer inspection. The synthesis model revisits these
passages in their source context and supplements the skill with supported
conditions and methodological relationships that were incompletely
expressed. The completion process preserves the initial skill and adds
guidance grounded in the source studies.

We evaluate the extracted knowledge across ten topics through conditional PPL, paired assessments of research plans, and answer-quality scores (Figure~\ref{fig:system}B). Directly synthesized skills lower PPL on both construction and held-out papers, and PPL-guided completion yields further reductions. To assess how skill guidance improves research planning, we compare plans generated without skills with those generated using directly synthesized (Direct) or refined (Refined) skills as references. In paired LLM evaluations, Direct-guided plans are preferred over no-skill plans in 87.9\% of comparisons, while Refined-guided plans are preferred over Direct-guided plans in 65.0\%. We further assess whether skill guidance helps the system address the methodological requirements of specific research tasks. For the same tasks, we compare responses generated without skills and with Direct or Refined skills as references. Each response is evaluated against paper-specific binary criteria, with its score reflecting the percentage of requirements satisfied. The mean percentage of requirements satisfied increases from 51.9\% without skills to 73.0\% with Direct skills and 74.5\% with Refined skills.

A selected drug-repurposing example illustrates how this guidance affects
specific scientific choices (Figure~\ref{fig:paper2skill-repurposing}). In
planning the investigation of AZD3355 for NASH-associated fibrosis,
skill-guided refinement makes the primary disease contrast and drug-profile
selection rule more explicit, and distinguishes prevention from treatment
of established fibrosis in the proposed validation experiments. These
changes illustrate how reusable methodological knowledge can help an agent
formulate a research plan with clearer connections between the scientific
objective, analytical choices, and experimental design.

\begin{figure}[p]
\centering
\begingroup
\setlength{\parskip}{0pt}
\setlength{\parindent}{0pt}
\renewcommand{\baselinestretch}{1}
\fontfamily{ptm}\selectfont
\definecolor{scTeal}{HTML}{087E83}
\definecolor{scPale}{HTML}{EDF6F5}
\definecolor{scInk}{HTML}{182C35}
\definecolor{scMuted}{HTML}{53666B}
\definecolor{scQuote}{HTML}{F4F5F6}
\definecolor{scLine}{HTML}{D6DFE1}
\definecolor{scRefinedFill}{HTML}{EAF6F4}
\definecolor{scRefinedLine}{HTML}{A4D1CB}
\tcbset{scCard/.style={enhanced,colback=scQuote,colframe=scLine,boxrule=.4pt,
  arc=1mm,left=2.3mm,right=2.3mm,top=1.3mm,bottom=1.3mm,
  before skip=0pt,after skip=0pt,
  fontupper=\fontsize{8.7}{11.2}\selectfont\RaggedRight}}
\tcbset{scRefined/.style={colback=scRefinedFill,colframe=scRefinedLine,
  overlay={\draw[->,>=stealth,line width=.7pt,draw=scTeal!75]
    ([xshift=-3.8mm]frame.west)--([xshift=-.9mm]frame.west);}}}
\tcbset{minimum for equal height group=nashSkills:106.06859pt,
  minimum for equal height group=nashAnswer0:85.39981pt,
  minimum for equal height group=nashAnswer1:60.18906pt,
  minimum for equal height group=nashAnswer2:58.28563pt}
\newcommand{\scsection}[2]{\par\vspace{1.3mm}\noindent
  {\sffamily\bfseries\color{scTeal}\fontsize{9.3}{12}\selectfont
  \tikz[baseline=-.6ex]\node[circle,fill=scTeal,text=white,inner sep=2pt,font=\bfseries\footnotesize]{#1};\hspace{2mm}#2}\par\vspace{1.7mm}}
\newcommand{\sclabel}[2][scTeal]{{\sffamily\bfseries\color{#1}\fontsize{8.8}{10}\selectfont #2}\par\vspace{1.5mm}}
\begin{tcolorbox}[enhanced,colback=scPale!25!white,colframe=scTeal,boxrule=.65pt,
title={Example: skill-guided decisions in drug repurposing},colbacktitle=scTeal,coltitle=white,
fonttitle=\sffamily\bfseries\fontsize{10.5}{14}\selectfont,
before skip=0pt,after skip=0pt,arc=1mm,left=3.5mm,right=3.5mm,top=2.5mm,bottom=2.5mm,
fontupper=\fontsize{8.7}{11.2}\selectfont\RaggedRight,colupper=scInk]
\textbf{Research task.} Assess whether AZD3355 (lesogaberan), a previously evaluated GABA-B receptor agonist, could be repurposed to limit or reverse fibrosis and disease progression in non-alcoholic steatohepatitis (NASH).
\scsection{A}{Skill evolution: from a shared workflow to conditional guidance}
\noindent\begin{minipage}[t]{.485\linewidth}\vspace{0pt}
\begin{tcolorbox}[scCard,equal height group=nashSkills]
\sclabel[scMuted]{INITIAL SKILL}
Define the disease contrast; construct signed expression signatures; match perturbation profiles to the relevant biological context; prioritize candidates; validate in increasingly relevant models.
\par\smallskip
Keep distinct disease stages separate. Include controls for nonspecific toxicity and define advancement criteria.
\end{tcolorbox}
\end{minipage}
\hfill
\noindent\begin{minipage}[t]{.485\linewidth}\vspace{0pt}
\begin{tcolorbox}[scCard,scRefined,equal height group=nashSkills]
\sclabel{REFINED SKILL: ADDED GUIDANCE}
With multiple drug profiles, prespecify selection or summarize consistency across contexts rather than choosing the strongest score.
\par\smallskip
Check that validation models represent the queried biology. Combine orthogonal phenotypes with confirmation of the predicted expression change before advancement.
\end{tcolorbox}
\end{minipage}
\scsection{B}{Research decisions: more explicit choices at three key steps}
\noindent\begin{minipage}[t]{.485\linewidth}\sclabel[scMuted]{ANSWER WITH INITIAL SKILL}\end{minipage}\hfill\begin{minipage}[t]{.485\linewidth}\sclabel{ANSWER WITH REFINED SKILL}\end{minipage}
\noindent\begin{minipage}[t]{.485\linewidth}\vspace{0pt}
\begin{tcolorbox}[scCard,equal height group=nashAnswer0]
\sclabel[scMuted]{Research contrast}
``The primary contrast is human NASH liver versus non-NASH control liver''
\par\smallskip
``NASH with advanced fibrosis (F3--F4) versus mild/no fibrosis (F0--F2).''
\end{tcolorbox}
\end{minipage}
\hfill
\noindent\begin{minipage}[t]{.485\linewidth}\vspace{0pt}
\begin{tcolorbox}[scCard,scRefined,equal height group=nashAnswer0]
\sclabel{Research contrast}
``\textbf{Primary contrast.} Human liver biopsy NASH with advanced fibrosis (F3--F4) versus NASH with no/mild fibrosis (F0--F1).''
\par\smallskip
``This isolates fibrosis/progression biology more directly than a broad NASH-vs-healthy contrast.''
\end{tcolorbox}
\end{minipage}
\par\vspace{1.7mm}
\noindent\begin{minipage}[t]{.485\linewidth}\vspace{0pt}
\begin{tcolorbox}[scCard,equal height group=nashAnswer1]
\sclabel[scMuted]{Drug-profile selection}
``extract profiles from liver-relevant cell lines and multiple time points/doses.''
\end{tcolorbox}
\end{minipage}
\hfill
\noindent\begin{minipage}[t]{.485\linewidth}\vspace{0pt}
\begin{tcolorbox}[scCard,scRefined,equal height group=nashAnswer1]
\sclabel{Drug-profile selection}
``Do not simply select the strongest-scoring profile; \textbf{prespecify a selection rule} or summarize consistency across cell types, doses, and times.''
\end{tcolorbox}
\end{minipage}
\par\vspace{1.7mm}
\noindent\begin{minipage}[t]{.485\linewidth}\vspace{0pt}
\begin{tcolorbox}[scCard,equal height group=nashAnswer2]
\sclabel[scMuted]{Treatment timing}
``Use a validated diet-induced or diet-plus-chemical NASH fibrosis model that reproduces fibrosis and metabolic disease.''
\end{tcolorbox}
\end{minipage}
\hfill
\noindent\begin{minipage}[t]{.485\linewidth}\vspace{0pt}
\begin{tcolorbox}[scCard,scRefined,equal height group=nashAnswer2]
\sclabel{Treatment timing}
``Design prevention and reversal arms: start AZD3355 \textbf{before fibrosis or after established fibrosis}.''
\end{tcolorbox}
\end{minipage}
\scsection{C}{More explicit decisions for treatment-focused validation}
The refined answer prioritizes the fibrosis contrast, specifies how drug profiles are selected, and separates prevention from treatment of established disease.
\par\vspace{2mm}\begingroup\renewcommand{\arraystretch}{1.12}\centering
\begin{tabular*}{.94\linewidth}{@{}l@{\extracolsep{\fill}}rr@{}}\toprule
Condition & Mean answer score $\uparrow$ & Reference-methodology PPL $\downarrow$\\
\midrule
No skill & 73.3\% & 10.521\\
Initial skill & 80.0\% & 10.008\\
Refined skill & 93.3\% & 9.987\\
\bottomrule\end{tabular*}\par\endgroup
\end{tcolorbox}
\endgroup

\vspace{2mm}
\caption{\textbf{Skill-guided research planning for AZD3355 in NASH-associated fibrosis.} \textbf{A,} Adapted excerpts illustrating the initial workflow guidance and conditional rules added during skill refinement. \textbf{B,} Example comparisons of answers before and after skill refinement. \textbf{C,} Comparison of no-skill, initial-skill, and refined-skill conditions. Mean answer score is the percentage of five task-specific evaluation criteria satisfied, averaged across three complete answers per condition. PPL is calculated on the reference methodology extracted from the source paper~\cite{bhattacharya2021azd3355}, conditioned on the research task and the corresponding skill when provided.}
\label{fig:paper2skill-repurposing}
\end{figure}

\subsection{Self-Scaling Discovery}
\label{subsec:ad}

Self-Scaling Discovery enables \methodabbr{} to investigate a broad space of
hypotheses from a single research question. Given an objective and starting
evidence, the agent builds a set of candidates, investigates them in parallel,
and decides what to pursue next.

\begin{figure}[!t]
    \centering
    \includegraphics[width=\linewidth]{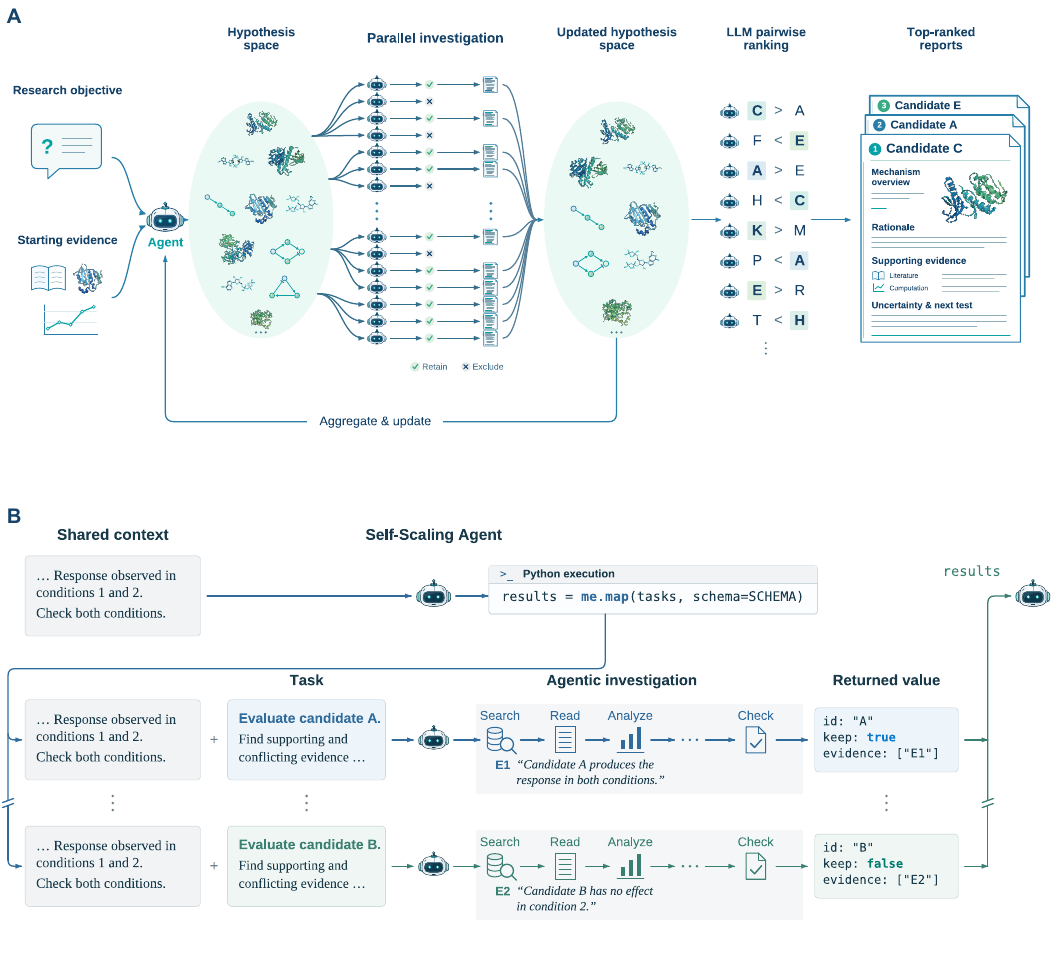}
    \caption{\textbf{Self-Scaling Discovery: hypothesis-space exploration and
    Self-Scaling Agent execution.}
    \textbf{A,} Given a research objective and starting evidence, \methodabbr{}
    investigates hypotheses in parallel, retains supported candidates,
    updates the hypothesis space, and compares candidate reports through
    pairwise ranking. Top-ranked reports summarize mechanisms, rationale,
    supporting evidence, and questions for follow-up.
    \textbf{B,} The Self-Scaling Agent scales itself across candidates, shown
    here by calling itself from code it writes and executes. Each instance
    inherits the caller's full research context, receives an individual task,
    and conducts a sequence of research actions.}
    \label{fig:auto_discovery}
\end{figure}

Much of research is shaped by the cost of investigating an idea. Reading the
relevant literature, running analyses and working through conflicting evidence
all take time. Scientists use experience and intuition to choose a few
promising hypotheses early, often before they have enough evidence to assess
the alternatives. With cheaper and more capable AI, research can instead begin
with a much larger hypothesis space and use investigation itself to decide
where to focus, so that directions once set aside for lack of time receive
serious consideration.

In a target discovery problem, for example, the agent can start with a broad
collection of proteins and investigate how each might explain an observed
effect. An initial pass establishes what is known about the candidates.
Subsequent rounds follow the evidence, examining promising mechanisms in detail
or using computation to resolve an open question. A finding about one candidate
can prompt a new question about another. The agent brings these findings
together, updates the candidate space and decides how to organize the next
round. Once the investigation has narrowed the space, the agent compares the
remaining candidates pairwise and ranks them for experimental follow-up
(Figure~\ref{fig:auto_discovery}A).

Carrying this process across thousands of candidates exceeds what a single
agent can handle. Subagents extend its reach, but they receive the judgment
accumulated during an investigation only through their prompts, and an
inadequate prompt can bias their assessments. The Self-Scaling Agent instead
extends its own understanding of the problem and its research capabilities
across thousands of tasks, producing results that can be compared and combined
directly (Figure~\ref{fig:auto_discovery}B). Its dedicated tools build this
judgment and these capabilities into their operation; a tool that ranks
candidates, for example, assesses them as the agent currently would, and its
assessments become better informed as the investigation proceeds. From the code
it writes, the agent can also call itself, extending its judgment and
capabilities in more flexible ways.

In the study described in Section~\ref{subsec:wet_ad}, Self-Scaling Discovery
started from a single mechanistic question, investigated a space of 2,607
proteins and ranked the 452 candidates that remained through 1,808 pairwise
comparisons; the top-ranked target was subsequently confirmed by SPR to bind
the molecule. When intelligence is no longer scarce, Self-Scaling Discovery can
make systematic exploration of the hypothesis space the starting point for
scientific discovery.

\FloatBarrier
\section{PharmAgents-Bench v0.1}
\label{sec:benchmark}

PharmAgents-Bench v0.1 evaluates autonomous drug discovery at two
complementary levels: the quality of individual scientific decisions and
the ability to organize a complete investigation supporting candidate
delivery. The verifiable decision-making track examines bounded research
questions with reference outcomes derived from published evidence. The
end-to-end process evaluation track assesses how a system develops and
executes a research strategy and justifies its proposed candidates.

\subsection{Verifiable Decision Making}
\label{subsec:decision_benchmark}

Advancing a drug discovery project requires sound scientific decisions
about how to interpret evidence and what to pursue next. We therefore
designed the verifiable decision-making track to assess whether a system
can make appropriate decisions at individual research decision points,
given the project's objective and constraints. The track comprises 121
questions across nine drug discovery scenarios. Each question presents
a specific research situation and requires an explicit decision or
assessment that can be checked against reference evidence from published
papers and patents. Table~\ref{tab:decision_metrics} summarizes the
decision scenarios and their evaluation metrics.

\begin{table}[!htbp]
\centering\small
\renewcommand{\arraystretch}{1.18}
\setlength{\tabcolsep}{4pt}
\caption{Drug discovery decision scenarios and evaluation metrics.}
\label{tab:decision_metrics}
\begin{tabularx}{\linewidth}{@{}>{\raggedright\arraybackslash}p{0.30\linewidth}>{\raggedright\arraybackslash}X>{\raggedright\arraybackslash}p{0.23\linewidth}@{}}
\toprule
Task & Decision scenario & Evaluation metric \\
\midrule
Binding-site assessment
& Locate a binding pocket to guide structure-based ligand design.
& Residue accuracy \\
Target hypothesis prioritization
& Prioritize target hypotheses for a compound with an unexplained phenotype.
& Known-target hit@50 \\
Molecular edit prioritization
& Choose where and how to modify a parent structure during optimization.
& Top-1 accuracy \\
Development candidate selection
& Compare candidate molecules as a whole and select one to advance into development.
& Selection accuracy \\
Molecular optimization outcome prediction
& Anticipate property trade-offs during optimization toward a development candidate.
& Property-direction score \\
Experimental affinity ranking
& Prioritize candidates for testing by expected affinity.
& Pairwise ordering accuracy \\
Agonist/antagonist assessment
& Assess whether candidates produce the desired receptor response under specified conditions.
& Composite assessment score \\
Experimental evidence assessment: literature
& Assess whether published studies directly tested a compound in the specified disease context.
& Known-anchor hit@10 \\
Experimental evidence assessment: patents
& Assess whether patent disclosures report direct tests of a compound in the specified disease context.
& Known-anchor hit@10 \\
\bottomrule
\end{tabularx}
\end{table}

\begin{figure}[!ht]
\centering
\includegraphics[width=\linewidth]{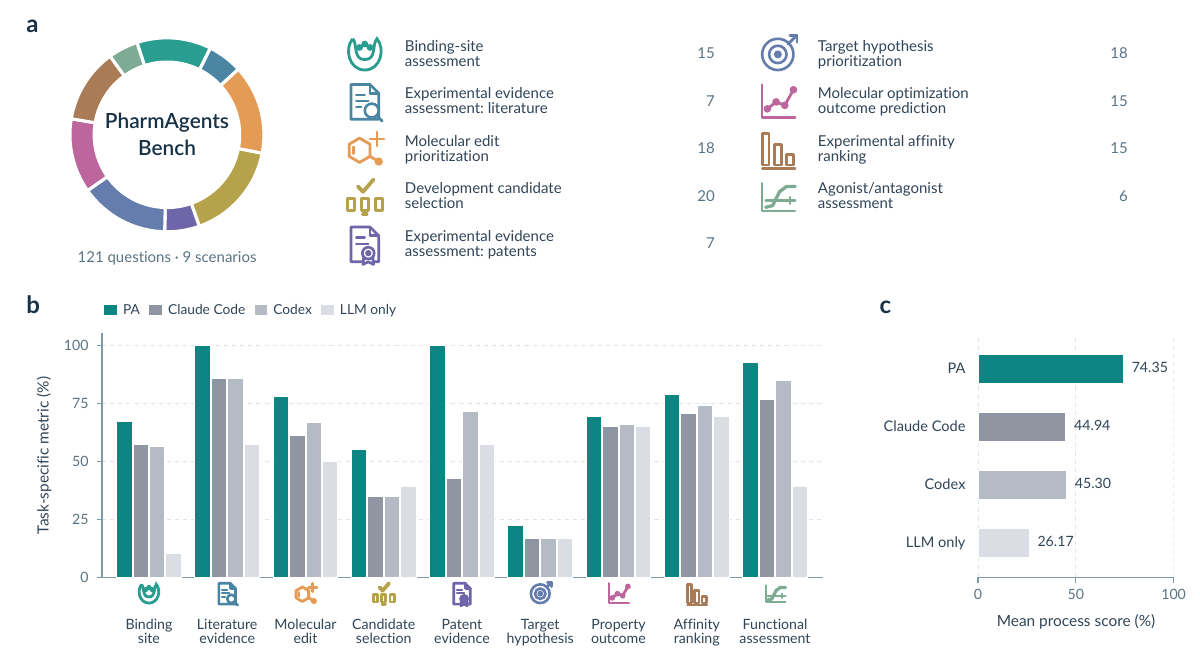}
\caption{\textbf{PharmAgents-Bench evaluates scientific decision-making and end-to-end research execution.}
\textbf{a}, Distribution of 121 questions across nine drug discovery decision
scenarios, spanning evidence assessment, hypothesis prioritization, and
molecular design and selection.
\textbf{b}, Performance of PharmAgents2 (PA), Claude Code, Codex, and the
standalone language model (LLM only), all using DeepSeek V4.1 Flash as the
base model. Results are expressed as percentages using the task-specific
metrics defined in Table~\ref{tab:decision_metrics}.
\textbf{c}, Research-process quality across three end-to-end drug discovery
tasks. An independent judge LLM evaluates recorded research activities,
computational artifacts, and final reports against expert-defined criteria
covering research strategy, method selection, computational investigation,
evidence interpretation, and candidate nomination. Scores are averaged
over three repeated runs per system and task, normalized to percentages,
and then averaged with equal weight across the three tasks.}
\label{fig:benchmark_controlled}
\end{figure}

A development-candidate-selection task illustrates this decision setting
by asking the system to choose among four closely related BRAF(V600E)
inhibitors for oral treatment of solid tumours with brain metastases.
Given the molecular structures and requirements for brain penetration,
high unbound brain exposure, and low efflux, the system must select one
candidate and justify its choice using permitted computational analyses,
without literature or external information retrieval. The reference is
Mol-A, corresponding to the reported development candidate
PF-07284890, whose identity is withheld
from the task input. In the recorded runs, PA combines structural analysis
with ADMET predictions and selects Mol-A, matching the reference, whereas
Claude Code and Codex emphasize molecular descriptors and select Mol-B.
The case illustrates how systems use different evidence to prioritize
candidates under the same project requirements.

\FloatBarrier
\begin{figure}[!htbp]
\begin{tcolorbox}[
  enhanced,colback=aipherMist,colframe=aipherTeal,
  boxrule=0.6pt,arc=0.8mm,
  left=3mm,right=3mm,top=2mm,bottom=2mm,
  colbacktitle=aipherTeal,coltitle=white,
  fonttitle=\sffamily\bfseries\small,
  title={Example: selecting a BRAF candidate for brain metastases},
  fontupper=\small]
\textbf{Agent-visible input.} Four BRAF(V600E) candidates rendered from the original SMILES.

\smallskip
\noindent\includegraphics[width=\linewidth]{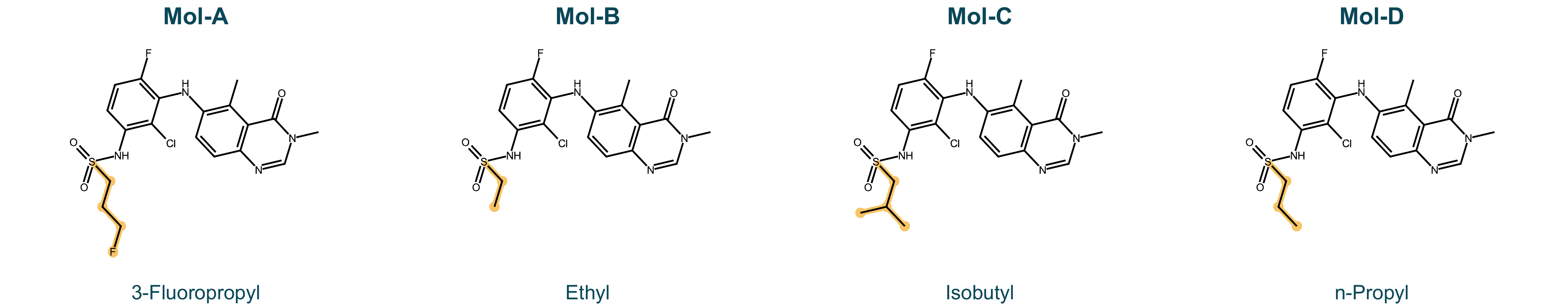}\par
\smallskip
\textbf{Project background:} ``The indication is BRAF-mutant solid
tumours / brain metastases. Oral; brain penetration is mandatory,
requiring a high Kp,uu and low efflux.''

\textbf{Decision:} Select one clinical candidate. Summarize tools used and
results; end with \texttt{Final Choice: Mol-X}. The original task prohibits
papers and external information retrieval.

{\centering
\includegraphics[width=\linewidth]{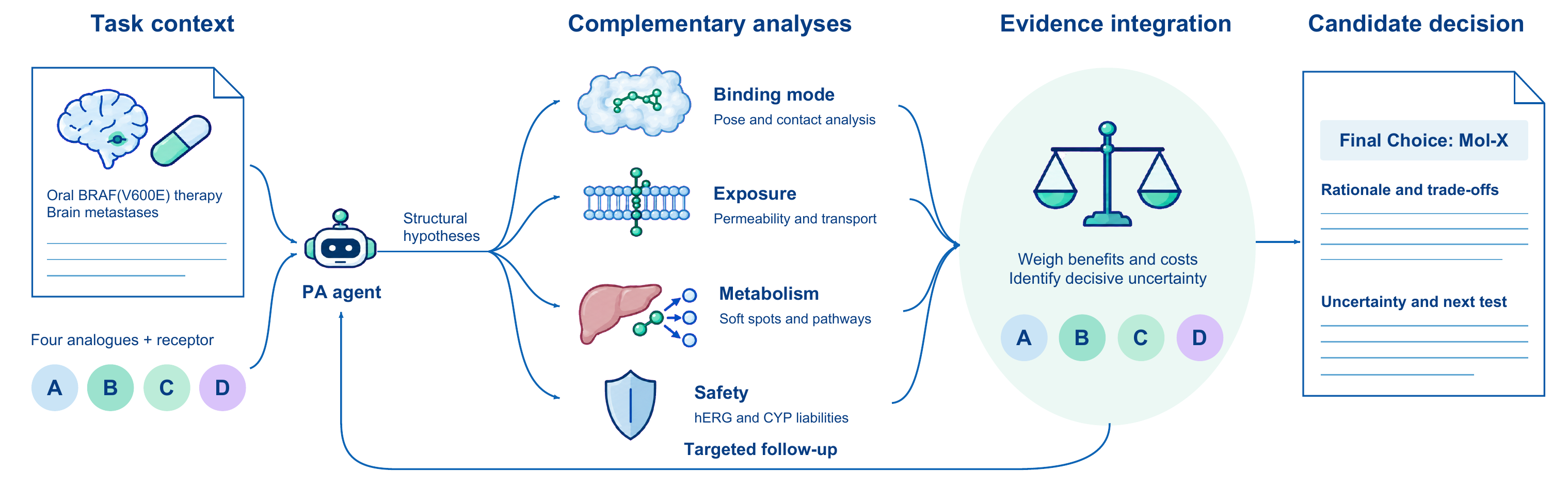}
\par}
\captionsetup{font=footnotesize,skip=3pt}
\caption{\textbf{Adaptive assessment in the BRAF development-candidate-selection
task.} The executed workflow integrates binding, exposure, metabolism, and
safety evidence to guide targeted follow-up and candidate selection.
Recorded final choices and supporting rationales for PA and the baseline
systems are summarized below.}
\label{fig:braf_decision_case}

\tcblower
\textbf{Observed outputs on the original question}
\smallskip

\begin{tabularx}{\linewidth}{@{}lXl@{}}
\toprule
System & Recorded final output & Reference agreement \\
\midrule
PharmAgents2 & \texttt{Final Choice: Mol-A} & \textcolor{aipherDarkGreen}{Match} \\
Claude Code & \texttt{Final Choice: Mol-B} & Different choice \\
Codex & \texttt{Final Choice: Mol-B} & Different choice \\
\bottomrule
\end{tabularx}

\smallskip
\textbf{PA's stated rationale:} BBB and metabolic predictions favoured A
over D, with brain exposure still uncertain. \textbf{Baseline rationales:}
Both favoured B's lower molecular weight, cLogP, and rotatable-bond count.

\smallskip
\textbf{Held-out reference: Mol-A (PF-07284890).} This source identity
was not included in the agent input.

\end{tcolorbox}
\end{figure}

We compare PharmAgents2 with Claude Code, Codex, and a standalone language
model across the nine decision-making task categories, using DeepSeek as
the base model in the reported configurations. Comparisons reflect each
system's configured access to prompts, scientific tools, and decision
skills. PharmAgents2 outperforms the baseline systems in this evaluation
(Figure~\ref{fig:benchmark_controlled}b), with gains in evidence assessment,
binding-site assessment, molecular edit prioritization, and affinity ranking.

\subsection{End-to-End Process Evaluation}
\label{subsec:e2e_benchmark}

The end-to-end delivery track evaluates whether a system can organize a
complete scientific investigation to support candidate delivery. It
comprises three tasks drawn from internal discovery projects: NK2R peptide
agonist design, JAK1 inhibitor discovery, and LpxH small-molecule design.
Each task specifies a discovery objective and project requirements, asking
the system to develop a research strategy and deliver candidate molecules
with supporting research reports.

Computational benchmarks cannot establish the experimental quality of the
delivered candidates, and scores from computational oracles are not
equivalent to the quality of the final discovery outcome. The research
process nevertheless admits scientifically grounded distinctions in
quality: strategies can address project requirements more completely,
methods can better fit the scientific questions, and conclusions can have
stronger evidential support.

Experts in medicinal chemistry and biology define task-specific scoring
rubrics covering research strategy, method selection, computational
investigation, evidence interpretation, and candidate nomination. These
criteria accommodate different valid approaches while assessing how well
an investigation addresses the project requirements and supports its
conclusions. PharmAgents2 and the baseline systems conduct their
investigations autonomously, and a judge LLM independently evaluates their
research processes and final reports against these criteria. PharmAgents2
outperforms the baseline systems in this evaluation
(Figure~\ref{fig:benchmark_controlled}c), indicating closer alignment with
the criteria defined by human experts.

\FloatBarrier
\section{Results: Wet-Lab Hypotheses Validation}
\label{sec:exp}

\subsection{Overview}
\label{subsec:results_overview}

Following the benchmark evaluation, we examine the deliverables of autonomous research in three
drug discovery studies and the experimental evidence supporting them. Researchers provided
discovery objectives and relevant starting information. \methodabbr{} autonomously formulated
research strategies, organized and executed computational investigations, and delivered candidate
lists with supporting research reports.

In the LpxH and JAK1 studies, researchers purchased and tested the nominated compounds. These experiments identified \textbf{AIPher-L1}, supported by enzyme inhibition and biophysical binding
assays, and \textbf{AIPher-J1}, which showed activity in two cytokine-signaling assays. A third
study applied Self-Scaling Discovery to secondary-target exploration for the anonymized molecule \textbf{AIPher-X} and delivered a
ranked list of candidate targets. Target X ranked first, and subsequent SPR
experiments supported binding. Together, the studies connect autonomous research to candidate delivery and
experimental evidence, as summarized in Table~\ref{tab:case_overview}.

\begin{table}[!htbp]
\centering\small
\caption{Overview of autonomous research, candidate delivery, and experimental evidence.}
\label{tab:case_overview}
\begin{tabularx}{\linewidth}{@{}p{0.15\linewidth}p{0.32\linewidth}X@{}}
\toprule
Case & Discovery narrative & Evidence status \\
\midrule
LpxH & Allosteric Site Exploration and Functional Assessment $\rightarrow$ Agent-Designed Discovery Plan $\rightarrow$ Candidate Molecules & 63 candidates purchased and tested; \textbf{AIPher-L1} identified with supporting evidence from enzyme inhibition and biophysical binding assays. Binding site identification is ongoing, with preliminary evidence supporting the proposed mechanism. \\
JAK1 & Known Covalent Site $\rightarrow$ Binding-Pocket Exploration $\rightarrow$ Candidate Molecules & 34 candidates purchased and tested; \textbf{AIPher-J1} exhibited inhibitory activity comparable to that of the positive control. Three additional compounds achieved greater than 50\% inhibition in both functional pathways. \\
Secondary-target discovery & Secondary-Target Hypothesis $\rightarrow$ Parallel Target Exploration $\rightarrow$ Ranked Target List and Research Reports & Exploration across 2,607 proteins produced 452 candidate dossiers and three target nominations, with Target X ranked first. Subsequent SPR experiments support binding of \textbf{AIPher-X} to Target X ($K_\mathrm{d}=4.64\,\mu\mathrm{M}$). \\
\bottomrule
\end{tabularx}
\end{table}

\FloatBarrier
\subsection{LpxH: Autonomous Research Planning and Hit Discovery}
\label{subsec:wet_lpxh}

The LpxH case (Figure~\ref{fig:lpxh_discovery}a) examines how \methodabbr{} translates an open-ended
antibacterial discovery objective into a research direction, a computational plan, and an
experimental handoff. The scientific motivation was the limited diversity of established LpxH
inhibitor scaffolds and binding modes. Published inhibitor development has largely centered on two
chemotypes: sulfonylpiperazine derivatives originating from AZ1~\citep{cho2020lpxh} and the acetylpiperazine series
identified by Zhang et al. through virtual screening in 2026~\citep{zhang2026lpxh}. Both retain a central
piperazine ring and primarily occupy the established L-shaped pocket associated with substrate
recognition. This concentration within related chemical and structural space motivated the
exploration of distinct scaffolds and alternative modes of inhibition, with the aim of broadening
antibacterial discovery opportunities and potentially addressing resistance.

Against this background, human researchers specified LpxH as the target and the discovery of
candidate inhibitors with novel scaffolds as the objective. They left the binding pocket, proposed
mechanism, and screening protocol open to investigation. \methodabbr{} integrated literature, known
ligand structures, chemical-space information, the LpxH structure, and computational pocket analysis
to define a discovery strategy. The analysis identified Site 1 as the established L-shaped pocket
targeted by inhibitors that compete with the substrate, and Site 2 as an alternative, potentially
allosteric pocket. \methodabbr{} prioritized Site 2, directing the search beyond the binding region
explored by established inhibitor classes.

Selection of Site 2 raised a further question: how might ligand binding at this location affect LpxH
function? \methodabbr{} examined functional, structural, and protein-interface evidence in relation
to the selected pocket and formulated three working hypotheses: (i) ligand-induced changes in helix
conformation could perturb the geometry of the Mn$^{2+}$-dependent catalytic center~\citep{young2013lpxh}; (ii) binding at
a putative dimer interface~\citep{huseby2024lpxh} could disrupt intersubunit interactions and interfere with dimer
formation; or (iii) the ligand could act as a molecular glue, stabilizing a nonproductive dimer
conformation and thereby impairing enzyme function. These three putative allosteric mechanisms
acting outside the substrate-binding pocket provided a rationale for pursuing Site 2 through
candidate screening and assessment, while its precise location and functional effects remain to be
validated.

The allosteric site hypothesis did not determine a fixed screening pipeline. The next research
decision was how to make it actionable when no known ligands or activity measurements for the
proposed pocket were available as a task-specific reference. According to the project account,
\methodabbr{} first planned a GenPack step to prioritize pseudo-ligands and select their corresponding
pockets for DrugCLIP virtual screening. It then selected the TargetMol in-stock library (100,000+ compounds) from the
platform's available resources and organized a candidate-discovery strategy.
DrugCLIP supported broad retrieval against the selected pocket, and DrugLens supported further
affinity ranking; further physicochemical analysis, molecular docking, pose and interaction
interpretation addressed complementary uncertainties raised by the evolving candidate set. In this
organization, intermediate outputs were inputs to the next decision---which candidates required
additional analysis, which form of evidence was needed, and how candidate priorities should be
updated---rather than independent scores in a predetermined sequence.

During execution, \methodabbr{} repeatedly integrated the available evidence for each candidate.
Broad retrieval and ranking identified molecules that merited closer examination; predicted poses
and interactions helped assess the structural plausibility of their proposed binding modes; and
physicochemical properties informed their priority for testing, particularly the balance between
lipophilicity and hydrophilicity, which could influence penetration across the Gram-negative
bacterial outer membrane and thereby antibacterial activity. The system used this combined
assessment to advance, deprioritize, or further analyze candidates and to determine when the
evidence was sufficient for experimental nomination. The resulting parallel deliverables were a list
of 63 candidate molecules and an accompanying research report recording the investigation,
rationale, supporting evidence, and remaining uncertainty. The nominated molecules, rather than the
report itself, then proceeded to human-executed testing.

Human researchers purchased and tested 63 candidates, identifying \textbf{AIPher-L1}
(Figure~\ref{fig:lpxh_discovery}b), a small molecule with a molecular weight of 282 Da, with
supporting evidence from enzyme inhibition and biophysical binding assays. \textbf{AIPher-L1}
inhibited KpLpxH with an IC$_{50}$ of 8.38 $\mu$M and exhibited binding affinities of 1.40 $\mu$M by
SPR and 0.87 $\mu$M by MST. Its inhibitory activity remained basically unchanged across the tested
substrate concentrations, providing preliminary evidence consistent with the proposed allosteric
mechanism. These findings support continued investigation of the three working hypotheses described
above, while the binding site and specific mechanism remain to be established. \textbf{AIPher-L1}
also exhibited low structural similarity to the published LpxH inhibitors included in the
comparison, suggesting a distinct scaffold for further exploration. MICs of 32 $\mu$g/mL against
efflux pump deficient ($\Delta$TolC) \textit{K. pneumoniae} and \textit{E. coli} strains provided
preliminary evidence of antibacterial activity under these conditions.

The case connects site exploration and functional investigation to research planning, integrated
candidate assessment, and experimental handoff. The computational plan is itself an agent-produced
research output, with the candidate list and report providing the final deliverables. Human-executed
experiments support the delivered hit. The discovery objective of scaffold novelty and the proposed
site mechanism require their own evidence; this system-level account does not isolate the
contribution of a particular skill or computational tool.

\FloatBarrier
\subsection{JAK1: Selectivity-Guided Candidate Delivery for a Known Target}
\label{subsec:wet_jak1}

The second case focused on selective JAK1 inhibition as a strategy to suppress cytokine signaling
implicated in inflammatory and autoimmune diseases while sparing other JAK isoforms, particularly
JAK2, whose inhibition can disrupt hematopoiesis. This approach offers the potential to preserve
anti-inflammatory efficacy while reducing adverse effects associated with broader JAK inhibition.
During its literature review, \methodabbr{} found that established approaches to JAK1 selectivity,
particularly those exploiting interactions with loop regions around the ATP pocket, had been
extensively explored. To pursue a distinct discovery strategy, the system therefore investigated
covalent targeting of cysteine residues as an alternative route to JAK1 selectivity.

In reviewing the literature against the discovery objective outlined above, \methodabbr{} identified
three key findings that informed its research strategy. First, Kavanagh et al. demonstrated that
covalent engagement of Cys817 in the JAK1 pseudokinase domain by VVD-118313 selectively inhibited
JAK1 transphosphorylation and cytokine signaling, with chemical proteomic profiling and C817A
mutagenesis supporting its site of action. Although these compounds also engaged the corresponding
cysteine in TYK2, they largely spared TYK2 signaling~\citep{kavanagh2022jak1}. Second, structural analysis by
Caveney et al. proposed that VVD-118313 binding at Cys817 perturbs the linker between the
pseudokinase (PK) and tyrosine kinase (TK) domains, destabilizing the arrangement required for
productive transactivation within the JAK dimer and thereby impairing transphosphorylation and
downstream signaling~\citep{caveney2023jak}. Third, a subsequent study showed that natamycin covalently
engaged JAK1 Cys988 outside the ATP pocket and suppressed cellular JAK1 signaling in a
Cys988-dependent manner without inhibiting recombinant JAK1 catalytic activity~\citep{natamycin2025jak1}.
Together, these findings provided \methodabbr{} with an experimental and structural rationale for
exploring covalent inhibition of JAK1.

\methodabbr{} further identified Cys817 as a promising starting point because its covalent
engagement and functional consequences had been systematically characterized, providing a coherent
experimental foundation for further discovery. At the same time, reported pharmacokinetic
limitations of existing ligands highlighted opportunities for optimization, including metabolic
stability. \methodabbr{} therefore proposed exploring alternative scaffolds for
covalent targeting of Cys817, with JAK1 selectivity and properties relevant to further development
as discovery objectives. This assessment led to an open-ended research question: could new chemical
scaffolds support covalent engagement of JAK1 Cys817 while offering opportunities to improve
selective functional inhibition and metabolic stability? The resulting task specified a known target
residue while leaving the covalent warhead, scaffold, surrounding binding pocket, and candidate
selection strategy open to investigation.

\setlength{\textfloatsep}{12pt plus 2pt minus 2pt}
\begin{figure}[!p]
\centering
\includegraphics[width=\linewidth,height=0.93\textheight,keepaspectratio]{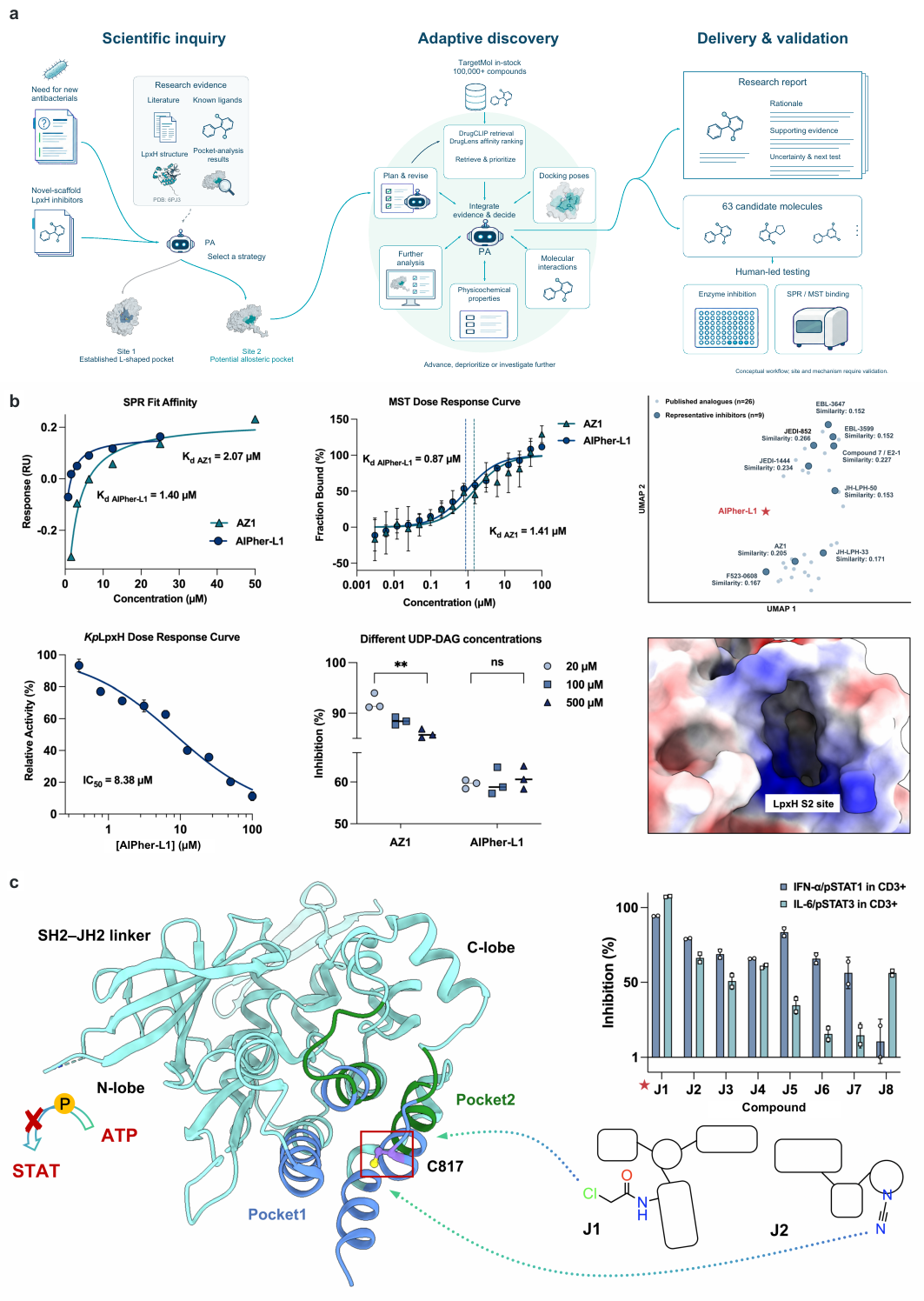}
\caption{\textbf{Autonomous research planning and candidate discovery for LpxH and JAK1. a}, Conceptual reconstruction of the LpxH investigation.}
\label{fig:wet_cases}
\label{fig:lpxh_discovery}
\label{fig:jak1_delivery}
\par\hfill{\footnotesize\itshape\color{aipherMuted}(continued on next page)}
\end{figure}

\begin{figure}[!t]
\raggedright\small
\setlength{\parskip}{0pt}
\noindent{\color{aipherNavy}\textbf{Figure~\ref{fig:lpxh_discovery} (continued)}}\quad
Literature, known ligands, structural information, and pocket analysis inform selection of the potentially allosteric Site 2 over the established Site 1 pocket. An agent-designed plan uses DrugCLIP for retrieval from the TargetMol in-stock library and DrugLens for affinity ranking. Integrated assessment of physicochemical properties, docking poses, and molecular interactions guides follow-up and candidate prioritization. The system delivers 63 candidate molecules and a parallel research report; human researchers test the molecules. The proposed site and mechanisms remain to be validated. \textbf{b}, Experimental characterization of \textbf{AIPher-L1} for LpxH, including SPR and MST binding measurements, concentration-dependent inhibition of KpLpxH, and inhibition at different substrate concentrations. Chemical-space analysis illustrates its structural distinction from published LpxH inhibitors, while the surface representation highlights the proposed Site 2 pocket. \textbf{c}, Structural rationale for covalent targeting of JAK1 Cys817, showing two candidate pockets and warhead schematic structures of representative compounds \textbf{AIPher-J1} and \textbf{AIPher-J2}. Functional assays assess candidate inhibition of IFN-$\alpha$/pSTAT1 and IL-6/pSTAT3 signaling in CD3$^{+}$ cells at 10 $\mu$M. \textbf{AIPher-J1} is highlighted for its inhibitory activity comparable to that of the positive controls BMS-986165 and Tofacitinib in both assays.
\end{figure}

Considering Cys817 lies in a relatively open region of the JAK1 JH2 pseudokinase domain,
\methodabbr{} first investigated neighboring cavities to identify pockets suitable for candidate
screening. This analysis identified two candidate pockets around Cys817. GenPack was applied to each
pocket, and the resulting pseudo-ligands were prioritized according to their predicted poses,
spatial compatibility, and proximity to Cys817, with particular attention to geometries potentially
compatible with covalent bond formation. The selected pocket representations were then used for
DrugCLIP virtual screening of the TargetMol in-stock covalent compound library (3500+). In parallel,
\methodabbr{} analyzed patents related to VVD-118313, summarized the reported structure--activity
relationships, and consolidated these insights into a reusable Skill to guide candidate assessment.
DrugLens provided further ranking by predicted affinity, followed by filtering based on
physicochemical properties and predicted ADME characteristics, with emphasis on metabolic stability,
bioavailability, and half-life. Covalent docking and comparative analysis of poses and
protein--ligand interactions informed the final selection of 34 compounds for purchase and
experimental testing.

Human researchers purchased and tested the 34 candidates in two functional assays measuring
IFN-$\alpha$-induced STAT1 phosphorylation and IL-6-induced STAT3 phosphorylation in CD3$^{+}$
cells. At 10 $\mu$M, \textbf{AIPher-J1} exhibited inhibitory activity comparable to that of the
positive controls, BMS-986165 and Tofacitinib, while three additional compounds achieved greater
than 50\% inhibition in both pathways (Figure~\ref{fig:lpxh_discovery}c). Prompted by its activity
in both functional assays, further investigation revealed that AIPher-J1 had previously been studied
in an oncology context. The present findings extend its functional characterization to cytokine
responses in immune cells, revealing a signaling phenotype beyond those examined in its previous
studies. \methodabbr{} autonomously identified two candidate druggable pockets around Cys817 and
used them to guide screening, yielding molecules with novel scaffolds and covalent warheads that
expand the chemical space for investigating potential JAK1 covalent inhibitors. Further experiments
are underway to validate JAK1 selectivity and the proposed binding modes.

\subsection{Self-Scaling Discovery: From a Known Molecule to Candidate Secondary Targets}
\label{subsec:wet_ad}

\begin{figure}[!htbp]
    \centering
    \includegraphics[width=\linewidth,height=0.83\textheight,keepaspectratio]{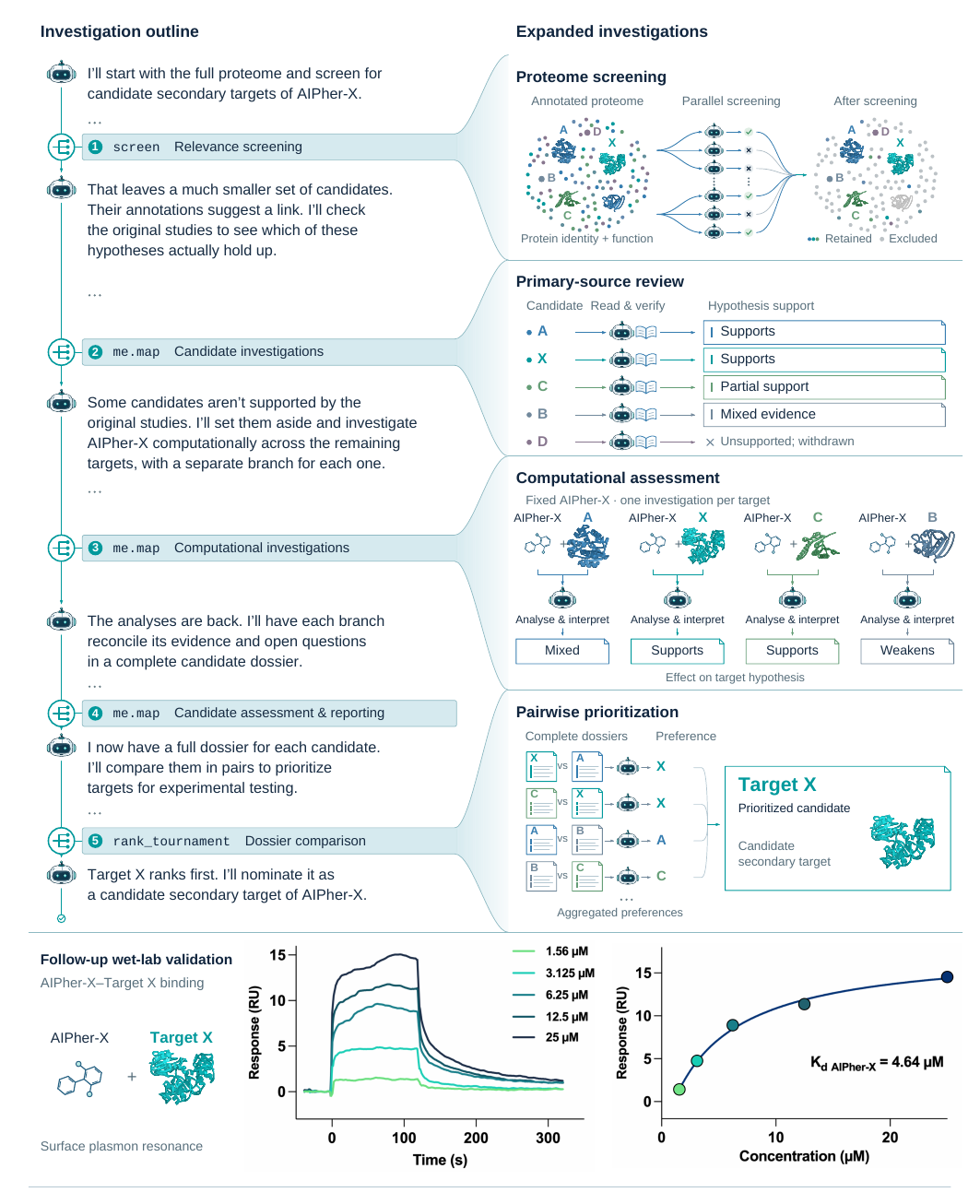}
    \caption{\textbf{Self-Scaling Discovery--guided secondary-target investigation for AIPher-X.}
    The schematic investigation outline on the left depicts five bulk tool calls that screened a
    reference proteome, investigated primary literature, conducted parallel
    computational assessments, prepared complete candidate dossiers, and ranked
    candidates through pairwise comparison. Expanded views on the right illustrate
    selected calls and how their findings inform candidate retention and
    prioritization. Complete dossiers integrated literature and computational
    evidence before entering the ranking stage. The investigation
    ranked Target X first. After PharmAgents2 completed the investigation, researchers
    performed wet-lab validation. The bottom row presents the resulting SPR sensorgrams and a concentration--response fit supporting
    \textbf{AIPher-X} binding to Target X, with a reported
    $K_\mathrm{d}$ of $4.64\,\mu\mathrm{M}$.
    The molecular structure shown for \textbf{AIPher-X} is for illustrative purposes only.
    Protein structures are substitute illustrations.}
    \label{fig:ad_target_discovery}
\end{figure}

A molecule with established biological activity can still have aspects of its mechanism left
unexplained. Researchers supplied the experimental background of \textbf{AIPher-X}, an anonymized
molecule: it binds a known primary target with high affinity, but inhibition of that target is
susceptible to resistance, whereas AIPher-X itself shows a low propensity for resistance. They
proposed that an additional target could help explain this difference and asked Self-Scaling Discovery to
investigate candidate secondary targets. Starting from this background and hypothesis, Self-Scaling Discovery
organized large-scale parallel exploration across a reference space of 2,607 proteins and produced
a ranked list of candidate targets
(Figure~\ref{fig:ad_target_discovery}).

The agent first combined protein annotations with the research background to formulate a specific
reason for investigating each candidate. Screening assessed whether perturbing the protein could affect
biological processes relevant to the molecule's activity, without requiring prior evidence that the
molecule binds the protein. This initial assessment advanced 1,719 candidates to literature
investigation. It turned a protein catalogue into a hypothesis space with concrete research questions,
opening the search to potential explanations beyond the known primary target.

Self-Scaling Discovery then investigated the candidates in parallel, tracing functional, genetic
perturbation, and pharmacological evidence to primary sources. Each branch examined why its
candidate merited investigation, what existing experiments supported, and which findings weakened
the initial hypothesis. Source-backed findings refined or replaced the starting annotations,
and this assessment retained 452 candidates for computational investigation of their potential
interactions with \textbf{AIPher-X}.

Computational investigations held the molecule fixed while examining different candidate targets.
Each branch organized analyses suited to its candidate, examining potential binding regions,
molecular interactions, and consistency with the available functional evidence. The branch
interpreted how these findings affected the target hypothesis. Literature and computational findings
were integrated into 452 complete candidate dossiers, each recording a proposed mode of
action, the principal evidence, conflicting findings, and experiments that could resolve the
remaining questions.

All 452 dossiers entered comparative assessment. Across eight rounds, 1,808 pairwise comparisons
evaluated their explanatory value, evidential support, and value for experimental testing. The
resulting ranked list nominated three targets for further investigation, with Target X ranked first.
The nominations were accompanied by reports explaining the selection and specifying the questions
to test.

After PharmAgents2 completed the investigation and delivered the ranked target list,
researchers conducted wet-lab validation of the top-ranked target. SPR measurements
showed binding between \textbf{AIPher-X} and Target X with a reported
$K_\mathrm{d}=4.64\,\mu\mathrm{M}$.

This case illustrates how Self-Scaling Discovery connects broad hypothesis exploration to concrete research
deliverables. Parallel investigations developed and assessed alternative target hypotheses, while
comparative assessment consolidated their findings into a ranked target list and supporting reports.
The completed investigation turned an open mechanistic question into prioritized targets for further
study, with subsequent wet-lab experiments supporting binding to the top-ranked candidate.

\section{Conclusion}
\label{sec:conclusion}

We formulate autonomous drug discovery as a continuous scientific decision-making process across a broad research space and develop PharmAgents2 to advance this process from research objectives to candidate delivery. Knowledge-enhanced decision making and organized large-scale exploration provide the two complementary capabilities underlying this approach. Benchmark evaluations demonstrate improved scientific decision quality and stronger execution of end-to-end investigations, while real-world studies yield new findings subsequently supported by experimental validation. These results establish a practical route for AI agents to organize and advance drug discovery, providing a prototype of a virtual pharmaceutical company.

A key next step is to use feedback from computational investigations and experimental validation to train and optimize the system. Successful and unsuccessful research outcomes can provide learning signals for improving scientific reasoning, refining decision knowledge, and guiding exploration. Connecting this feedback to agent training and skill refinement would allow experience gained within individual projects to strengthen capabilities across subsequent investigations. Such a continuous learning cycle offers a path toward virtual pharmaceutical companies that accumulate scientific knowledge and improve their research capabilities through experience, expanding the scale and scope of autonomous drug discovery.

\FloatBarrier
\begingroup
\titleformat*{\subsection}{\subsectionfont\sffamily\bfseries\color{black}}
\hypersetup{urlcolor=aipherMuted}

\section*{Contributors}
\label{sec:contributors}

\subsection*{Conceptualization}
Yanyan~Lan.

\subsection*{Project leads}
Haichuan~Tan (\email{thc23@mails.tsinghua.edu.cn})\\
Bowen~Gao (\email{billgao0111@gmail.com}).

\subsection*{Core contributors}
Tao~Huang, Jiaxin~Tan, Xin~Hong, Baoguo~Yao, and Yifei~Wang.

\subsection*{Contributors}
Zhichao~Deng, Hanji~Du, Dapeng~Jiang, Yiheng~Li, Weiwei~Ma, Xin~Ma,
Yuqi~Shi, Zixiang~Wang, Kelin~Wu, Tianxiang~Wu, Wenxuan~Xie,
Jie~Zang, Huiyu~Zhang, Mojia~Zhang, and Yuying~Zhang.

{\small\noindent$\ast$Last-Name in Alphabetical Order\par}

\subsection*{Supervisors}
Yanyan~Lan (\email{lanyanyan@tsinghua.edu.cn}), Ya-Qin~Zhang, and Wei-Ying~Ma.
\par
\endgroup

\ifshowbackmatter
\clearpage
\bibliographystyle{naturemag}
\bibliography{refs_nature}

\ifshowappendices
\clearpage

\appendix

\section{System Architecture}
\label{app:impl}

Figure~\ref{fig:system_architecture} shows how \methodabbr{} connects research
planning, decision support, and scientific computation. A task begins with
a research objective, project constraints, and available starting information.
The orchestration framework develops a plan and executes it step by step.
Within each step, the agent gathers evidence through tool calls, code
execution, and information retrieval. The resulting observations inform
subsequent actions and revisions to the remaining plan. Before delivery,
an independent reviewer examines the investigation and returns feedback
for revision when needed.

The decision core combines expert-authored and \paperoskill{}-derived skills
with evidence from scientific literature, patents, and computational analyses.
Skills guide method selection and interpretation in the context of the
current research question. A shared workspace retains data, code, and
intermediate results so that later steps can build on earlier findings.
The system delivers candidate recommendations together with a research
report documenting the supporting analyses and evidence.

AIPherDD provides access to in-house models, open-source models, and
scientific tools through a unified interface. The platform dispatches
computations to containerized runners, allocates CPU and GPU resources,
and monitors execution. Computational outputs return to the workspace
for further analysis and research decisions.

\begin{figure}[!htbp]
  \centering
  \includegraphics[width=\linewidth]{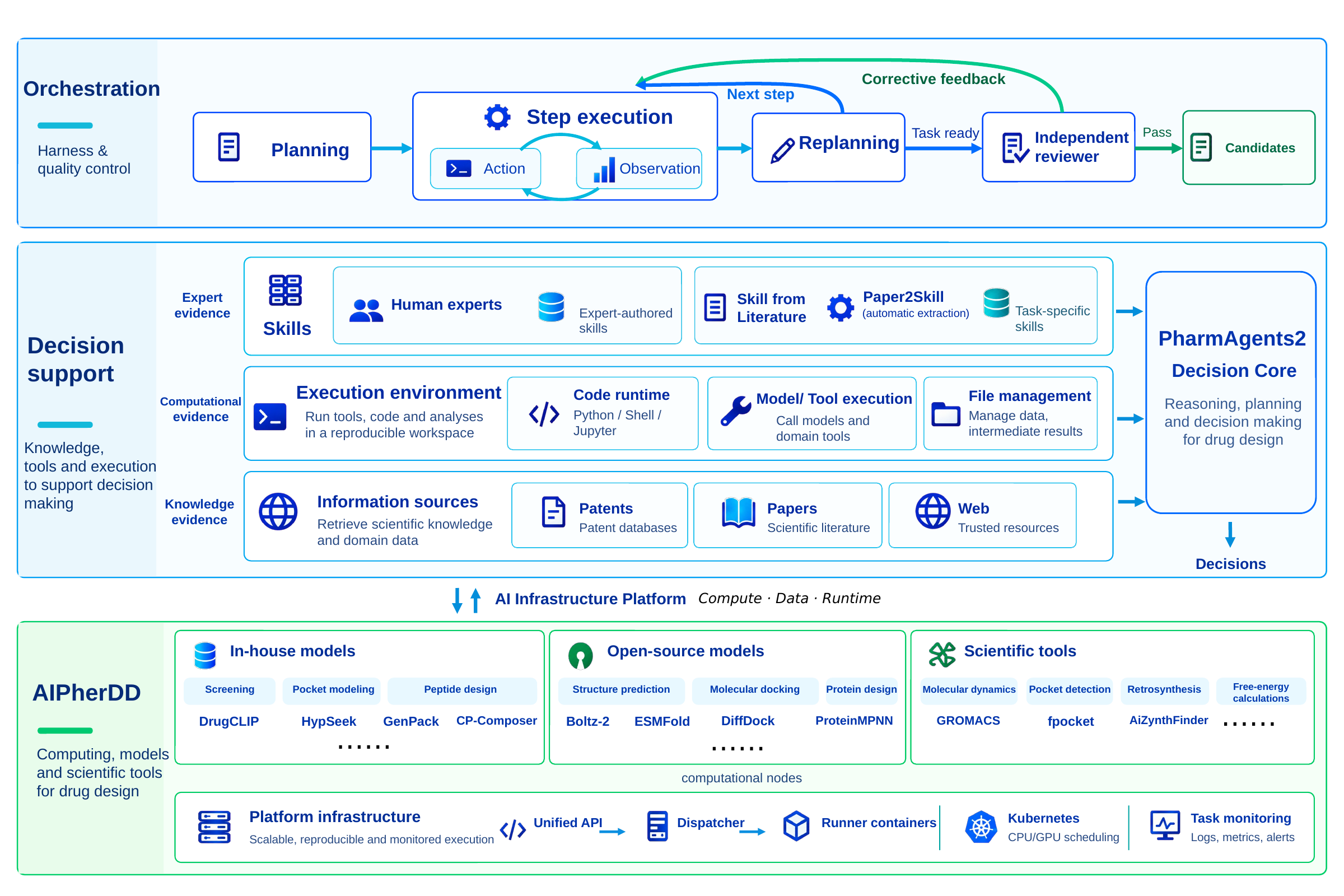}
  \caption{\textbf{System architecture and execution framework of \methodabbr{}.}
  The orchestration framework connects planning, step execution, replanning,
  and independent review. The decision core combines skills and research
  evidence within a shared execution environment. AIPherDD provides unified
  access to scientific models and tools, with distributed execution and
  resource management.}
  \label{fig:system_architecture}
\end{figure}

\clearpage

\section{Benchmark Details}\label{app:benchmark}
\label{app:decision_scenarios}

The nine task categories simulate research decisions made with incomplete
information during drug discovery. The descriptions below explain the
setting and decision represented by each category in
Table~\ref{tab:decision_metrics}.

\paragraph{Binding-site assessment.}
Before structure-based ligand design, a researcher must decide which region
of a protein to investigate for ligand binding. The task supplies a protein
structure with its ligand removed and asks the system to identify the
binding-site residues. This simulates locating the pocket that would guide
subsequent docking and molecular design.
\textit{Reference basis:} Residue labels are derived from the original
experimental complex using a 4.5~\AA{} heavy-atom contact cutoff to the
bound ligand, excluding crystallization additives.

\paragraph{Target hypothesis prioritization.}
A compound produces an experimental phenotype, but its direct molecular
target remains unresolved. Given the molecular structure and phenotype,
the system prioritizes specific proteins for target validation. The decision
requires connecting chemical and biological evidence while distinguishing
plausible direct targets from downstream pathway components.
\textit{Reference basis:} The reference protein is the target identified
in the source publication through direct binding or functional
target-deconvolution experiments.

\paragraph{Molecular edit prioritization.}
A medicinal chemistry team is refining a parent molecule and must decide
where and how to modify its structure. The system compares four proposed
edits, focusing on the structural changes relative to the parent and their
fit to the optimization goals. This simulates choosing a direction for
molecular modification.
\textit{Reference basis:} The label identifies the structural edit reported
in the source paper's development path, recording an observed design choice
without establishing it as the uniquely optimal modification.

\paragraph{Development candidate selection.}
A project has several candidate molecules and must decide which one to
advance into development. The system compares four existing candidates
as whole molecules, weighing their overall pharmacological and
developability profiles against the project requirements. This simulates
selecting the molecule to carry forward after comparing the available
candidates.
\textit{Reference basis:} The label is the clinical or preclinical candidate
nominated in the original discovery paper, recording the team's selection
without establishing that all alternatives are inferior.

\paragraph{Molecular optimization outcome prediction.}
A team must anticipate how a starting compound's property profile may change
as optimization proceeds toward a development candidate. Given the starting
structure and project context, the system predicts the direction of change
in specified endpoints, including trade-offs in which some properties are
sacrificed to improve others.
\textit{Reference basis:} Direction labels are calculated from published
experimental values for the starting compound and the development candidate
using predefined endpoint-specific thresholds.

\paragraph{Experimental affinity ranking.}
A researcher must prioritize compounds for testing against the same target
before their relative affinities are known. The system orders five candidates
from strongest to weakest expected affinity. This simulates allocating
follow-up effort based on comparative binding predictions.
\textit{Reference basis:} Pairwise orderings are derived from published
experimental affinity measurements for the candidates under the same assay
conditions.

\paragraph{Agonist/antagonist assessment.}
A project needs compounds that produce a particular receptor response, and
must decide which candidates warrant functional testing. The tasks combine
molecular comparisons, receptor structures, and assay context to assess
agonism or antagonism. They also examine whether functional conclusions
transfer across stereoisomers, receptors, experimental conditions, or
co-administration settings.
\textit{Reference basis:} Published functional experiments supply the hidden
response labels, while a predefined rubric separately evaluates reproducible
computation and pharmacological reasoning in the reported composite score.

\paragraph{Experimental evidence assessment: literature.}
Before investigating a compound in a specified disease context, a researcher must
establish what has already been tested. The system retrieves literature
describing direct experiments with the specified compound in that disease
setting. The decision concerns whether a source provides relevant
experimental evidence, including negative or null results, to inform
further investigation of the compound in that disease setting.
\textit{Reference basis:} Known source identifiers anchor published direct
experiments with the specified compound and disease context; hit@10 measures
recovery of these references rather than an exhaustive set of valid evidence.

\paragraph{Experimental evidence assessment: patents.}
Assessing a compound's experimental evidence may also require patent disclosures.
The system identifies experiments involving the specified compound within
the stated disease boundary. This simulates deciding which patent disclosures
provide usable experimental evidence, distinguishing performed studies
from broad claims or proposed experiments when planning subsequent research.
\textit{Reference basis:} Known patent identifiers anchor disclosed direct
experiments within the specified compound and disease scope; hit@10 measures
recovery of these references rather than an exhaustive set of valid evidence.

\fi
\fi

\end{document}